\documentclass[journal]{IEEEtran}
\usepackage{amsmath,amsfonts}
\usepackage{algorithmic}
\usepackage{algorithm}
\usepackage{array}
\usepackage[caption=false,font=normalsize,labelfont=sf,textfont=sf]{subfig}
\usepackage{textcomp}
\usepackage{stfloats}
\usepackage{url}
\usepackage{verbatim}
\usepackage{graphicx}
\usepackage{booktabs}
\usepackage[table,xcdraw]{xcolor} 
\usepackage{soul}                 
\sethlcolor{yellow}               
\usepackage[doipre={doi:~}]{uri}
\usepackage{booktabs,makecell, multirow, tabularx}
\usepackage[table,xcdraw]{xcolor}
\usepackage[colorlinks=true,
            linkcolor=blue,    
            citecolor=blue,    
            urlcolor=blue]{hyperref}

\begin{document}

\title{Airport Passenger Flow Forecasting via Deformable Temporal–Spectral Transformer Approach}

\author{
    Wenbo Du,~\IEEEmembership{Member,~IEEE}, 
    Lingling Han,~\IEEEmembership{Graduate Student Member,~IEEE}, 
    Ying Xiong,
    Ling Zhang,
   \\ Biyue Li,
    Yisheng Lv,~\IEEEmembership{Senior Member,~IEEE}, 
    Tong Guo*,~\IEEEmembership{Member,~IEEE}

\thanks{This work was funded in part by the National Natural Science Foundation of China (NSFC) under Grant U2333218 \textit{(Corresponding author: Tong Guo)}.}
\thanks{Wenbo Du, Lingling Han and Tong Guo are with the School of Electronic and Information Engineering and State Key Laboratory of CNS/ATM, Beihang University, Beijing 100191, China (e-mail: wenbodu@buaa.edu.cn; zy2402532@buaa.edu.cn; buaaguotong@foxmail.com).}
\thanks{Ying Xiong and Ling Zhang are with the Department of Information and Technology, Beijing Capital International Airport Co., Ltd., Beijing 100621, China (e-mail: xiongy@bcia.com.cn; zhangling3@bcia.com.cn).}
\thanks{Biyue Li is with the College of Information and Electrical Engineering, China Agricultural University, Beijing 100083, China (e-mail: libiyue@cau.edu.cn).}
 \thanks{Yisheng Lv is with the State Key Laboratory of Multimodal Artificial Intelligence Systems, Institute of Automation, Chinese Academy of Sciences, Beijing 100190, China, also with the Shandong Key Laboratory of Smart Transportation (Preparation), Shandong Jiaotong University, Jinan 250353, China, and also with the School of Artificial Intelligence, University of the Chinese Academy of Sciences, Beijing 100049, China (email: yisheng.lv@ia.ac.cn).}

}

\markboth{Journal of \LaTeX\ Class Files,~Vol.~14, No.~8, August~2021}%
{Shell \MakeLowercase{\textit{et al.}}: A Sample Article Using IEEEtran.cls for IEEE Journals}


\maketitle
\begin{abstract}
Accurate forecasting of passenger flows is critical for maintaining the efficiency and resilience of airport operations. Recent advances in patch-based Transformer models have shown strong potential in various time series forecasting tasks. However, most existing methods rely on fixed-size patch embedding, making it difficult to model the complex and heterogeneous patterns of airport passenger flows. To address this issue, this paper proposes a deformable temporal–spectral transformer (named DTSFormer) that integrates a multiscale deformable partitioning module and a joint temporal–spectral filtering module.
Specifically, the input sequence is dynamically partitioned into multiscale temporal patches via a novel window function-based masking, enabling the extraction of heterogeneous trends across different temporal stages. 
Then, within each scale, a frequency-domain attention mechanism is designed to capture both high- and low-frequency components, thereby emphasizing the volatility and periodicity inherent in airport passenger flows. 
Finally, the resulting multi-frequency features are subsequently fused in the time domain to jointly model short-term fluctuations and long-term trends. Comprehensive experiments are conducted on real-world passenger flow data collected at Beijing Capital International Airport from January 2023 to March 2024. The results indicate that the proposed method consistently outperforms state-of-the-art forecasting models across different prediction horizons.
Further analysis shows that the deformable partitioning module aligns patch lengths with dominant periods and heterogeneous trends, enabling superior capture of sudden high-frequency fluctuations.
\end{abstract}


\section{Introduction}
\IEEEPARstart{A}{ir} transportation serves as a vital link connecting the world. In recent years, airport passenger traffic has experienced continuous growth. In 2024, China's civil transport airports handled 1.46 billion passengers, reaching unprecedented levels in all three key operational indicators: passenger throughput, cargo and mail throughput, and aircraft movements \cite{CAAC2025}. 
The rapid growth of civil aviation passenger traffic has intensified the imbalance between airport resources and transport demand, placing considerable pressure on both airport internal resource allocation and the surrounding transportation system\cite{cai2024multiairport}. Airport passenger flow forecasting represents one of the most effective ways to balance airport resources with passenger demand. Reliable forecasts enable airport managers to allocate resources in advance, optimize traffic dispatching, and better handle peak-hour pressures and emergencies. This not only reduces passenger waiting times and improves travel experiences but also lowers operating costs and enhances resource utilization.

Existing airport passenger flow prediction methods can be divided into four categories: simulation methods, statistical regression methods, classical machine learning methods, and temporal deep learning methods. Simulation methods establish mathematical models to represent passengers in airports and to predict waiting times and passenger flows across key processes, such as security checks, boarding, baggage handling, and other procedures \cite{yin2024simulation, i2014passenger, li2018network, Zhang2014Agent}. This type of method is transparent and can intuitively reflect passenger processes at the airport, but it relies on oversimplified assumptions and thus suffers from limited accuracy \cite{ndoh1993evaluation, kim2004estimating}. 
Statistical regression methods, such as the Auto-Regressive Integrated Moving Average (ARIMA) model and the Gray prediction model, fit time-dependent airport passenger flow data to capture temporal variations \cite{hu2023air, liu2017improving}. These approaches offer high computational efficiency and improved accuracy. However, since airport passenger flow time series are typically nonlinear and irregular rather than smooth, their performance remains limited.
The rapid development of machine learning (ML) techniques has led to their wide application in airport passenger flow forecasting. Various ML approaches, such as Backpropagation Neural Networks (BPNN) \cite{ghanbari2010comparison}, Support Vector Regression (SVR) \cite{xie2014short}, and Random Forests (RF), fit the relationships between influencing factors and passenger volumes, offering broad applicability and strong nonlinear fitting capability. However, these methods often fail to fully capture the temporal dependencies inherent in passenger flow data, which distinguishes time series forecasting from general regression tasks.

Temporal deep learning has emerged as a promising approach for modeling passenger flow dynamics, leveraging recurrent neural networks (RNNs) to capture sequential patterns and temporal dependencies more accurately \cite{liu2017novel, brun2022predicting, xu2022short}. Various architectures, such as Long Short-Term Memory networks (LSTM) \cite{Jing2021Short} and Gated Recurrent Units (GRU) \cite{yu2022short}, are widely applied with demonstrated advantages in capturing temporal dependencies. However, these models struggle to model long-term dependencies due to gradual information decay in memory cells and also suffer from limited parallelization. To address these shortcomings, Transformer architectures have become a popular framework for time series forecasting. By leveraging the attention mechanism, Transformers can directly model dependencies across arbitrary time points in a single step while fully parallelizing computations across sequences. Representative models such as Informer \cite{zhou2021informer}, Autoformer \cite{wu2021autoformer}, iTransformer \cite{liu2023itransformer}, PatchTST \cite{nie2022time}, and Pathformer \cite{chen2024pathformer} consistently achieve state-of-the-art forecasting performance.

Among Transformer-based forecasting models, patch-enhanced approaches have attracted growing attention and deliver strong accuracy gains \cite{dai2024periodicity, tang2025unlocking}. Compared with non-patch strategies, they markedly improve performance by enriching local semantic representations. For instance, Crossformer \cite{zhang2023crossformer} segments time series into patches and applies self-attention across both temporal and cross-variable dimensions, boosting predictive power. PatchTST \cite{nie2022time} similarly partitions sequences into patches and incorporates instance normalization within a Transformer framework, yielding substantial improvements and broad adoption. Extending this line, Pathformer \cite{chen2024pathformer} notes that fixed-length patches can disrupt temporal continuity and therefore adapts patch lengths via time-series decomposition, offering preliminary flexibility. 
However, directly applying state-of-the-art patch-enhanced Transformer models to airport passenger flow forecasting is less effective, as their fixed or manually tuned patch sizes constrain flexibility and hinder adaptation to \textit{trend heterogeneity} and pronounced \textit{high-frequency fluctuations}.

Specifically, airport passenger flow is strongly shaped by calendar effects, i.e., weekends, fixed public holidays, and peak tourism seasons \cite{photun2023forecasting, gunter2021forecasting}. It is simultaneously vulnerable to uncertain external shocks such as temporary traffic control measures \cite{sun2019nonlinear} and adverse weather \cite{wei2009passenger}. These factors induce trend heterogeneity, i.e., distributional shifts across days (weekday vs. weekend/holiday) and within a day (morning/evening peaks). Consequently, fixed-length patches are inadequate to capture such heterogeneous temporal patterns. 
Moreover, concentrated flight arrivals often produce abrupt surges in passenger counts, yielding pronounced high-frequency components. Standard Transformers exhibit bottlenecks in handling high-frequency signals and under-utilize the full frequency spectrum \cite{yi2024filternet, yue2025freeformer}, even though high-frequency information is crucial for accurate airport passenger flow forecasting.


To address these challenges, we propose a novel deformable temporal–spectral Transformer, termed DTSFormer, for precise airport passenger flow forecasting. First, DTSFormer adaptively partitions the input sequence into multi-scale temporal patches via window-function masking, enabling effective extraction of heterogeneous trends across different temporal stages. Next, within each scale, a frequency-domain attention mechanism extracts both high- and low-frequency components, emphasizing the volatility and periodicity of airport passenger flows. Finally, the extracted multi-frequency features are fused in the time domain, jointly modeling short-term fluctuations and long-term trends.

In a nutshell, the innovative contributions are summarized as follows:
\begin{enumerate}
\item We propose a joint temporal-spectral modeling framework that decomposes airport passenger flow dynamics into complementary representations: \textit{temporal modeling} captures periodic patterns and heterogeneous trends, while \textit{spectral modeling} processes high-frequency fluctuations. This dual-domain collaborative approach enables comprehensive extraction of both transient variations and persistent temporal dependencies.
\item We develop an adaptive deformable partitioning mechanism that adjusts the number, offsets, and sizes of patches based on the input series in a data-driven manner, improving the capture of heterogeneous temporal patterns across stages.
\item We design a frequency domain modeling approach that processes queries and keys in the frequency domain using a modulus-based correlation to more accurately identify inter-series frequency relationships, which proves more effective than standard linear projections.
\item We propose a selective temporal fusion module that filters and integrates frequency components with learned, component-wise importance, enabling differentiated temporal modeling and improved predictive performance.
\end{enumerate}

We conduct comprehensive experiments based on real passenger flow data from Beijing Capital International Airport (ZBAA), covering the period from January 2023 to March 2024. The results show that the proposed method outperforms existing mainstream models and demonstrates superior predictive performance under different conditions. 
Further analysis shows that the deformable partitioning module learns patch lengths as fractional multiples of the dominant period, enhancing periodic pattern extraction, while its adaptive positioning aligns with heterogeneous passenger flow trends. Moreover, under pronounced high-frequency fluctuations, the model more effectively captures sudden changes than baseline methods.

The remainder of this paper is organized as follows. Section II reviews the existing literature on the prediction of airport passenger flow andproblem formulation. Section III introduces the proposed deformable temporal–spectral transformer. Section IV presents the passenger flow prediction results on ZBAA data and the feature importance analysis. In Section V, we further analyze the contributions of the two proposed innovative modules. In Section VI, we conclude this study and provide further discussion.

\section{BACKGROUND}
\subsection{Related Work}
\textit{1) Passenger Flow Forecasting Methods:}
Existing airport passenger flow forecasting methods can be divided into four categories: simulation and probability statistics methods, statistical regression methods, classical machine learning methods, and temporal deep learning methods. The number of passengers at each stage of the airport security process was simulated using a simulation and probability statistics method, and measures to reduce passenger flow were studied \cite{takakuwa2003simulation}. With the advent of the big data era, the use of simulation and probabilistic statistical methods has become relatively limited. Considering the temporal characteristics of passenger flow, an increasing number of studies have employed statistical regression methods for prediction. Statistical regression methods such as ARIMA and its variants are widely applied in prediction tasks. For example, the application of the ARIMA model for short-term freeway traffic flow prediction was investigated \cite{ahmed1979analysis}. Further on, it was shown that railway passenger flows exhibit strong seasonal autocorrelation, and the seasonal ARIMA method can effectively fit and forecast such time series with good predictive performance \cite{milenkovic2018sarima}. Passenger flow mean and volatility have also been modeled using four generalized autoregressive conditional heteroscedasticity models in combination with ARIMA, which demonstrated superior accuracy and reliability compared to traditional methods \cite{chen2019subway}. With the continuous development of computational techniques, machine learning models have been proposed to overcome the limitations of traditional approaches in capturing nonlinear patterns. For example, a modified SVR-based approach for holiday passenger flow prediction was developed, integrating an improved particle swarm optimization algorithm for parameter tuning and a pruning algorithm for sparsity \cite{liu2017holiday}.

Gradually, temporal deep learning methods, which combine the advantages of machine learning and statistical regression methods, have received more attention \cite{zhou2024multimodal}. Over the past decade, RNN \cite{hihi1995hierarchical}, LSTM \cite{hochreiter1997long}, and GRU \cite{cho2014learning} have demonstrated remarkable capabilities in capturing temporal dependencies, inspiring numerous studies to employ time series models for passenger flow prediction \cite{ye2024survey}. A deep learning architecture that combines residual networks, graph convolutional networks, and LSTM was proposed for forecasting short-term passenger flow in urban rail transit at the network scale \cite{zhang2020deep}. An end-to-end deep learning framework, termed Deep Passenger Flow, was introduced to predict inbound and outbound passenger flow, enabling the joint modeling of external environmental factors, temporal dependencies, and spatial characteristics \cite{liu2019deeppf}. In addition, an LSTM–based model was employed to predict passenger flow at Paris Charles de Gaulle Airport, producing forecasts for different security areas \cite{monmousseau2020predicting2}. These findings collectively demonstrate that temporal deep learning methods achieve strong performance in passenger flow prediction tasks.

\textit{2) Transformer for Time Series Forecasting:}
The Transformer is the mainstream method for natural language processing tasks (NLP) \cite{vaswani2017attention}, relying on the self-attention mechanism to extract semantic relations, and has achieved excellent results in this task. In passenger flow prediction tasks, Transformer is also widely used \cite{lin2024deep,wang2025efficient}. A novel Transformer-based architecture, termed the Spatial-Temporal Transformer Network, was proposed under the encoder–decoder framework to capture complex and dynamic spatial dependencies through the use of multiple graphs in a self-adaptive manner \cite{zhang2023cov,shi2024adaptive}. To extract multi-scale information, a hybrid deep learning model was developed, consisting of two parallel blocks: a GRU neural network with an attention mechanism and an improved Transformer \cite{zhang2022short}. These studies highlight the advantages of Transformer-based models in passenger flow prediction.

ViT is the first work in computer vision tasks that completely using Transformer architecture, and it creatively proposed the idea of “patch”. The concept of splitting images into patches to achieve image recognition performance comparable to convolutional neural networks was first demonstrated in the Vision Transformer \cite{dosovitskiy2020image}, highlighting the advantages of patches in computer vision. Building on this idea, patch-based improvements to Transformer models have shown excellent performance across various domains \cite{huang2019ccnet}. More recently, the PatchTST model was proposed for multivariate time series prediction, introducing patching into time series analysis and leveraging patches to extract local semantic information for improved forecasting performance \cite{nie2022time}. However, because PatchTST relies on fixed-length patches, it requires manual adjustment of patch size to match data characteristics, which undermines the robustness of the model. Adapting fixed parameters dynamically can improve performance. In the field of computer vision, there has been a large amount of work focusing on the adaptive ability of models \cite{chen2021dpt}, \cite{dai2017deformable}, \cite{fu2017look}. However, adaptive algorithms are less commonly used in the field of time series prediction. Pathformer \cite{chen2024pathformer} applied a multi-scale Transformer model with adaptive pathways, realizing a preliminary adaptive function. However, the model only selects the better effective value from a few predefined patch lengths and does not achieve full adaptivity.

Inspired by the Patch-based Transformer and adaptive pathways, we consider its application in airport passenger flow forecasting. However, it is difficult to capture heterogeneous temporal patterns using only one scale of patch. Therefore, we propose a model with multi-scale deformable capabilities that can adaptively learn the patches at multiple scales depending on the data features. In our model, the numbers, positions, and lengths of patches are the result of adaptive learning, without predefining any parameters.

\subsection{Problem Formulation}
In this paper, we focus on predicting the total number of passengers departing the airport per unit time. Accurate prediction of departing passengers enables airport managers to allocate egress-related resources, such as taxis and subways, in advance. This facilitates the rational allocation of resources while enhancing operational efficiency and service quality. 

\begin{figure}[!t]
\centering
\includegraphics[width=9cm,height=7.3cm]{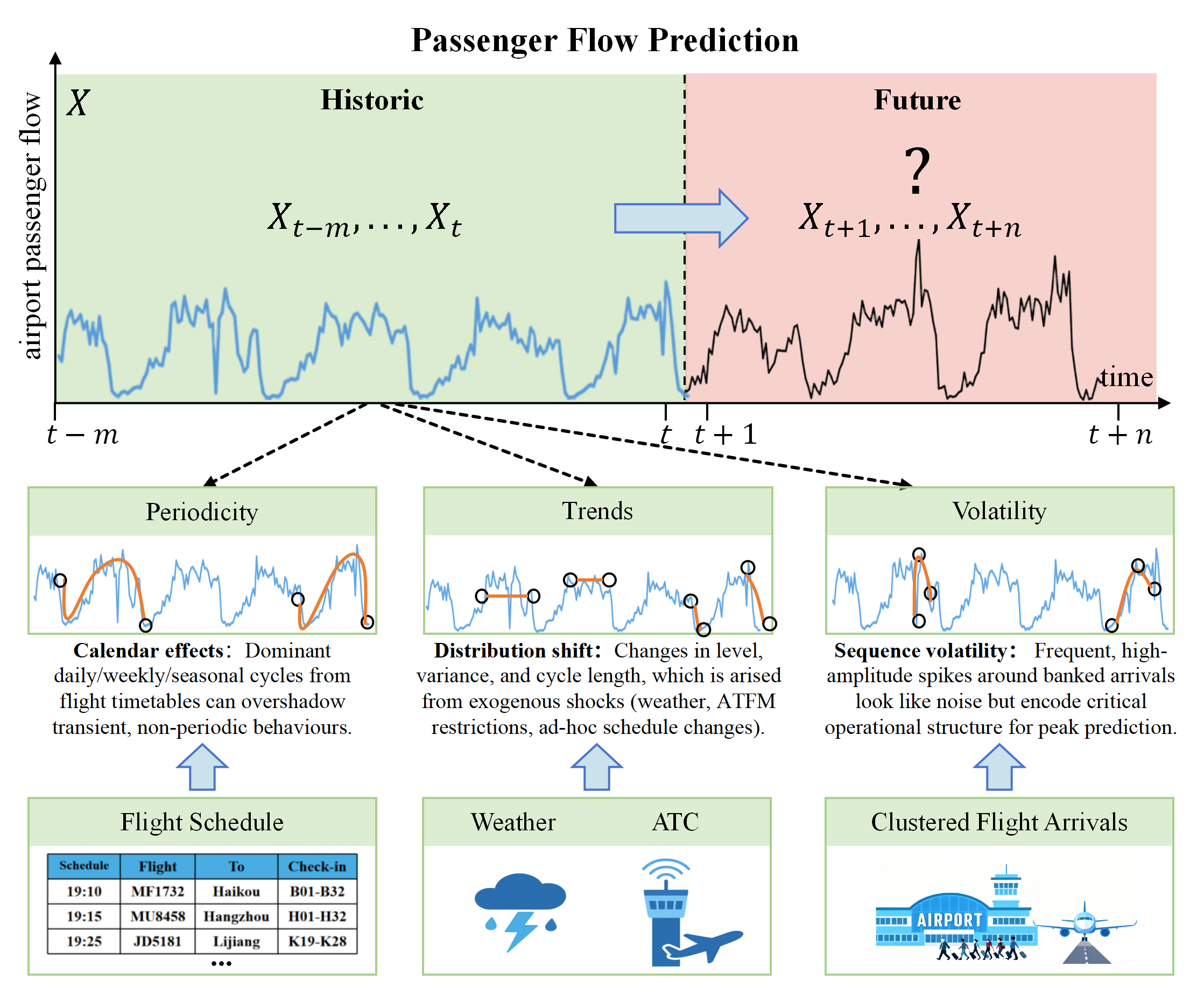}
\caption{An illustration of airport passenger flow that highlights its three salient characteristics: periodicity, heterogeneous trends, and high-frequency volatility, where ATC stands for air traffic control.}
\label{fig1}
\end{figure}

As shown in Fig.~\ref{fig1}, flight schedules—strongly tied to calendar effects (dates and holidays)—impose pronounced periodic patterns on passenger flow. Beyond this regularity, passenger flow is highly sensitive to uncertain factors such as temporary traffic controls, weather, and flight delays, leading to trend heterogeneity. Moreover, large passenger loads and clustered arrivals generate sharp short-term spikes, amplifying volatility, especially when these factors interact. In summary, airport passenger flow exhibits three salient traits, i.e., periodicity, trend heterogeneity, and significant volatility, whose coupling substantially increases the difficulty of accurate forecasting.

The airport passenger flow forecasting task can be formulated as a regression problem, where the prediction model provides estimates of the passenger flow for a given airport. Specifically, the model takes $m$ historical time steps as input and generates predictions for the subsequent $n$ time steps. This problem can be represented as:
\begin{align}
{\hat Y_{t + 1}},...,{\hat Y_{t + n}} = {\cal F}({X_{t - m}},...,{X_t};\theta )
\end{align}

\begin{align}
\mathop {\min }\limits_\theta  {\cal L}(\theta ) = \sum\nolimits_{t + 1}^{t + n} {L({{\hat Y}_t}(\theta ),{Y_t})}
\end{align}
where $m$ denotes the length of the historical time series and $n$ denotes the length of the forecasting time series, ${X_t}$ represents the flow of passengers at time $t$. ${\hat Y_t}(\theta )$ represents the predicted value of passenger flow at time $t$ and ${Y_t}$ denotes the true value. The sequence $({\hat{Y}}_{t+1}, \ldots, {\hat{Y}}_{t+n})$ represents the predicted passenger flow for the future period, computed by the function $\mathcal{F}$ based on the historical inputs $({X}_{t-m}, \ldots, {X}_t)$ and the learnable parameter vector $\theta$. The learning objective is to obtain the optimal parameter vector $\theta$ by minimizing the total loss $\mathcal{L}(\theta)$, which measures the discrepancy between the predicted and true passenger flows over all forecasted steps.

\section{the proposed method}
\subsection{Overview}
\begin{figure*}[!t]
\centering
\includegraphics[width=16cm,height=8cm]{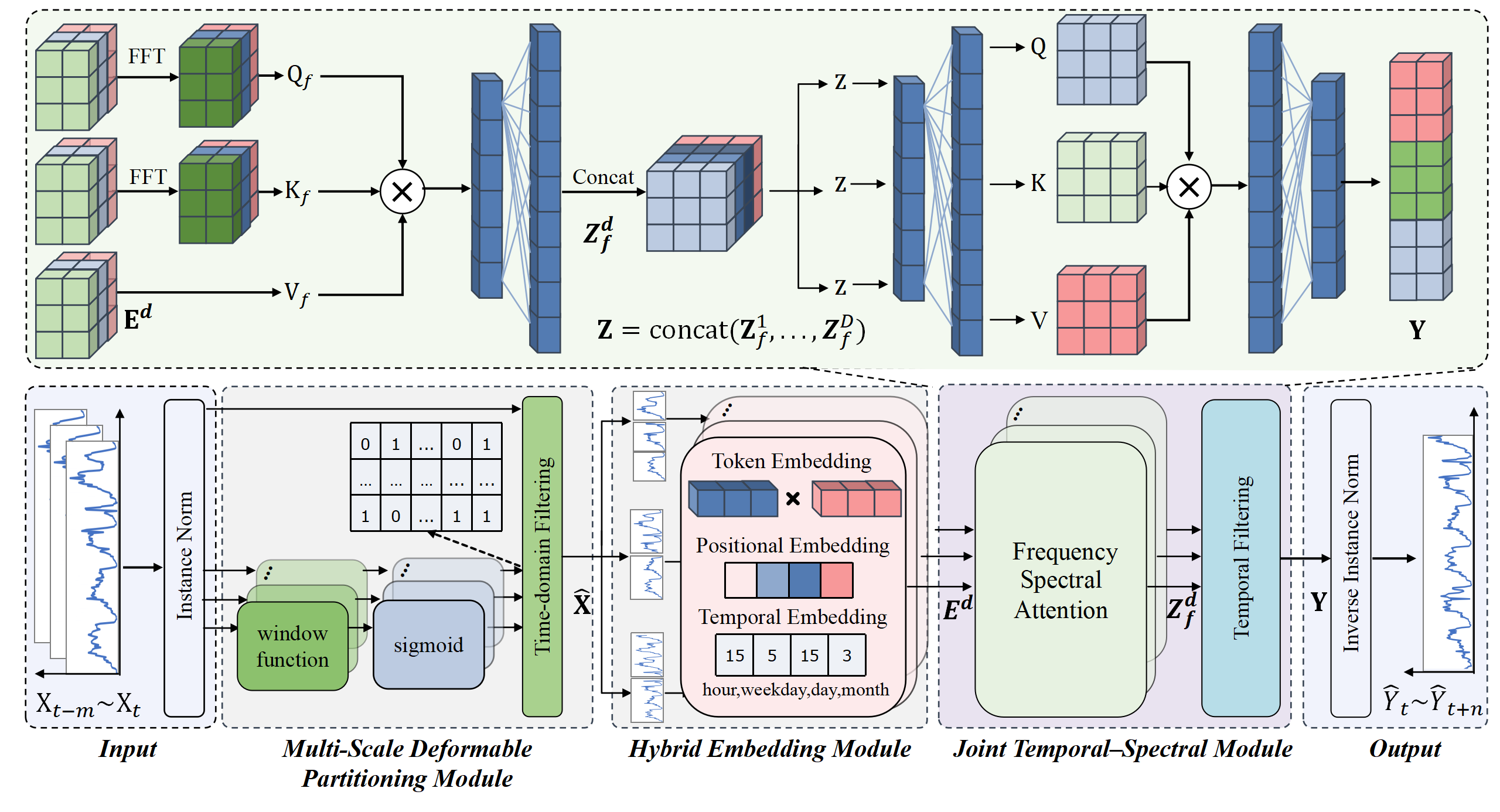}
\caption{Overall architecture of DTSFormer. The model consists of three key components: the multi-scale deformable partitioning module (left), the hybrid embedding module (middle), and the joint temporal–spectral filtering module (right).}
\label{fig2}
\end{figure*}

In this section, we propose a deformable temporal–spectral transformer, termed DTSFormer, designed for accurate passenger flow forecasting. The architecture of the proposed model is shown in Fig.~\ref{fig2}. The proposed architecture consists of three core components: a multi-scale deformable partitioning module that adaptively generates patches representations across different temporal scales; a hybrid embedding module that provides a more comprehensive characterization of passenger flow time series; and a joint temporal–spectral filtering module that integrates frequency-domain attention with time-domain fusion to enhance feature extraction.

To extract multi-scale temporal dependencies, DTSFormer employs multiple deformable layers. Each layer adaptively generates patches with variable numbers, offsets, and scales according to the characteristics of the input sequence, thereby capturing temporal patterns at different resolutions. This multi-scale design enhances the model’s ability to capture heterogeneous trends. Furthermore, at each scale, a joint temporal–spectral filtering mechanism is introduced to account for highly volatile patterns, enabling the model to capture both short-term fluctuations and long-term trends.

In the following sections, we describe the structure of the multi-scale deformable partitioning module, the hybrid embedding module, and the joint temporal–spectral filtering module.

\subsection{Multi-Scale Deformable Partitioning Module}
Airport passenger flow data are highly non-stationary, characterized by heterogeneous trends caused by external disturbances such as weather or flight delays. A fixed-length patch fails to adequately capture such heterogeneous temporal dynamics. To address this, the proposed module introduces a multi-scale deformable partitioning mechanism, which adaptively adjusts the number, size, and offsets of patches. The core idea is to align the partitioning process with the intrinsic structure of the data, thereby enabling the model to more effectively capture distributional shifts and heterogeneous patterns across different temporal stages.

\begin{figure*}[!t]
\centering
\includegraphics[width=17.5cm,height=5cm]{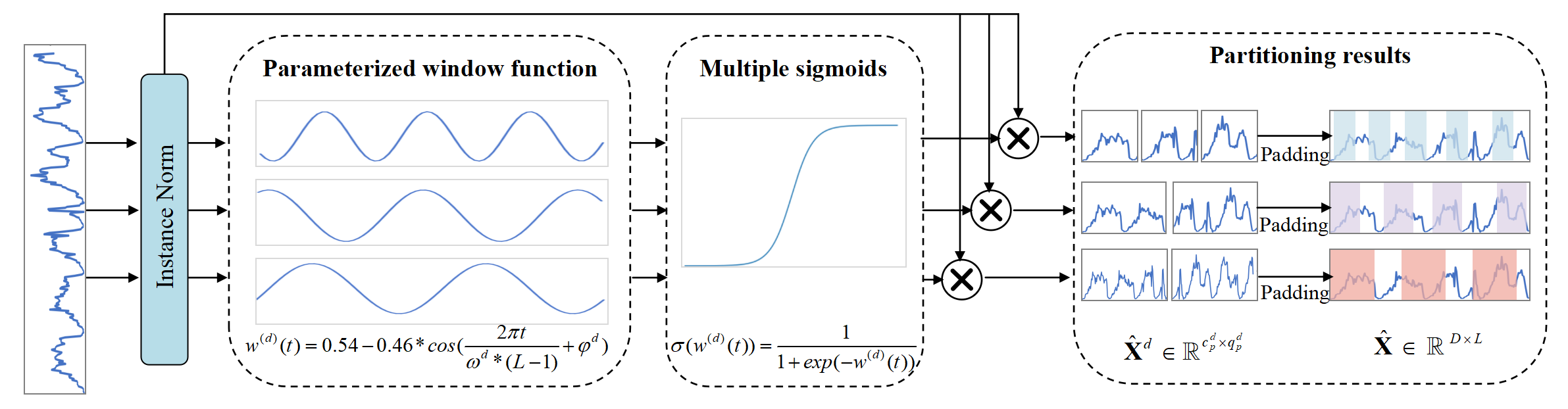}
\caption{The structure of the multi-scale deformable partitioning module. The input sequence is normalized and processed by window functions at different scales. A sigmoid-based gating mechanism adaptively generates masks, allowing the model to form deformable patches that capture heterogeneous temporal trends.}
\label{fig3}
\end{figure*}

We propose the multi-scale deformable partitioning module based on a window function, as illustrated in Fig.~\ref{fig3}. First, the input time series are transformed via instance normalization to remove scale effects and stabilize the data distribution. Then, a set of window functions with different scales is applied, generating multiple candidate temporal partitions that capture diverse periodic and trend characteristics. Next, these windowed signals are passed through a sigmoid-based gating mechanism with filtering, which adaptively adjusts each partition. Finally, the inputs are multiplied by the learned masks to form multi-scale deformable patches, enabling the model to flexibly capture heterogeneous temporal trends in passenger flow data.

The input ${\bf{X}} = [{\bf{X}}_{1:L}^1,{\bf{X}}_{1:L}^2, \ldots ,{\bf{X}}_{1:L}^D] \in {\mathbb{R}^{D \times L}}$ represents the passenger flow data, where $L$ denotes the length of the input sequence and $D$ denotes the data dimension. First, we utilize an instance normalization method \cite{kim2021reversible}, denoted as Norm, on input $\mathbf{X}$, which can be formulated as:
\begin{align}
\text{Norm}({\bf{X}}) = 
\left[
\frac{{{\bf{X}}_{1:L}^d - \text{mean}_L({\bf{X}}_{1:L}^d)}}{{\text{std}_L({\bf{X}}_{1:L}^d)}}
\right]_{d = 1}^D
\end{align}
where $mean{_L}()$ denotes the operation that calculates the mean value along the time dimension, and $std{_L}()$ represents the operation that calculates the standard deviation along the time dimension.

In the multi-scale deformable partitioning module, the operations at each scale are conducted independently and in parallel. The trainable parameters are distinct across different layers, allowing each scale to adaptively learn its own partitioning strategy. In each layer, we employ the Hamming window function to perform variable partitioning of the input sequence ${{\bf{X}}^d} \in {\mathbb{R}^{1 \times L}}$, where $d = 1,2,\ldots,D$ represents the $d$-th deformable module. The window function is adopted because its shape is characterized by a prominent center and smooth edges \cite{podder2014comparative}, which makes it suitable for use as a patch selector, retaining the data in the central region as effective patches. Based on the original definition of the Hamming window, where the coefficients 0.54 and 0.46 are retained as in the classical formulation, we introduce two additional trainable parameters to endow the Hamming window with learnable capability, ${\varphi ^d}$ and ${\omega ^d}$. Specifically, ${\omega ^d}$ controls the period and ${\varphi ^d}$ introduces a phase shift. The window function is defined as follows:

\begin{align}
w^{(d)}(t) = 0.54 - 0.46\cos\!\left(\frac{2\pi t}{\omega^{d}(L-1)} + \varphi^{d}\right)
\end{align}
where ${w^{(d)}}$(t) denotes the window function at the $d$-th scale, $0 \le t \le L - 1$; $t$ denotes the discrete position index within the window and $L$ represents the total length of the window. This formulation allows the window to adaptively emphasize selected regions while maintaining smooth transitions at the boundaries.

We apply a sigmoid function to the parameterized Hamming window to ensure differentiability and constrain its values between 0 and 1.
\begin{align}
{\tilde w^{(d)}}(t) = \sigma ({w^{(d)}}(t)) = \frac{1}{{1 + \exp( - {w^{(d)}}(t))}}
\end{align}

To extract the most relevant regions, patches are defined as the positions where the masked values exceed a threshold of $b$. This procedure provides a mechanism for selectively emphasizing informative segments of the input while preserving gradient flow for training.
\begin{align}
P_t^d = \begin{cases} 
1, & {\tilde w}^{(d)}(t) > b \\
0, & \text{otherwise}
\end{cases}, \quad 0 \le t \le L - 1
\end{align}
where $P_t^d$ denotes the binary patch mask at position $t$, with $P_t^d = 1$ indicating that the position is selected as a patch and $P_t^d = 0$ otherwise.

This operation emphasizes the time slots of interest in ${{\bf{X}}^d}$ while attenuating less relevant ones, allowing the model to focus on the most informative parts of the time series. This process can be formulated as:
\begin{align}
{{\bf{\hat X}}^d} = {{\bf{X}}^d}\odot{{\bf{P}}^d}
\end{align}
where ${{\bf{\hat X}}^d} \in {\mathbb{R}^{c_p^d \times q_p^d}}$ denotes the patch sequence obtained at the $d$-th scale, $c_p^d$ denotes the patch length, and $q_p^d$ denotes the number of patches. Both the patch length $c_p^d$ and the number of patches $q_p^d$ at each scale are not fixed but are instead adaptively learned from the data characteristics.

To ensure dimensional consistency, the data points that are not selected as patches are padded with zeros. Each represents the time series segmented under a specific scale, ${{\bf{\hat X}}^d}$ in which selected positions retain their original values, while unselected positions are set to zero. After processing through deformable layers, we obtain $\mathbf{\hat X} = [\mathbf{\hat X}^1, \ldots, \mathbf{\hat X}^d, \ldots, \mathbf{\hat X}^D] \in \mathbb{R}^{D \times L}$. This operation ensures that only the selected positions contribute to the subsequent processing, effectively partitioning the input into informative segments.

\subsection{Hybrid Embedding Module.}
\begin{figure}[!t]
\centering
\includegraphics[width=8.6cm,height=8cm]{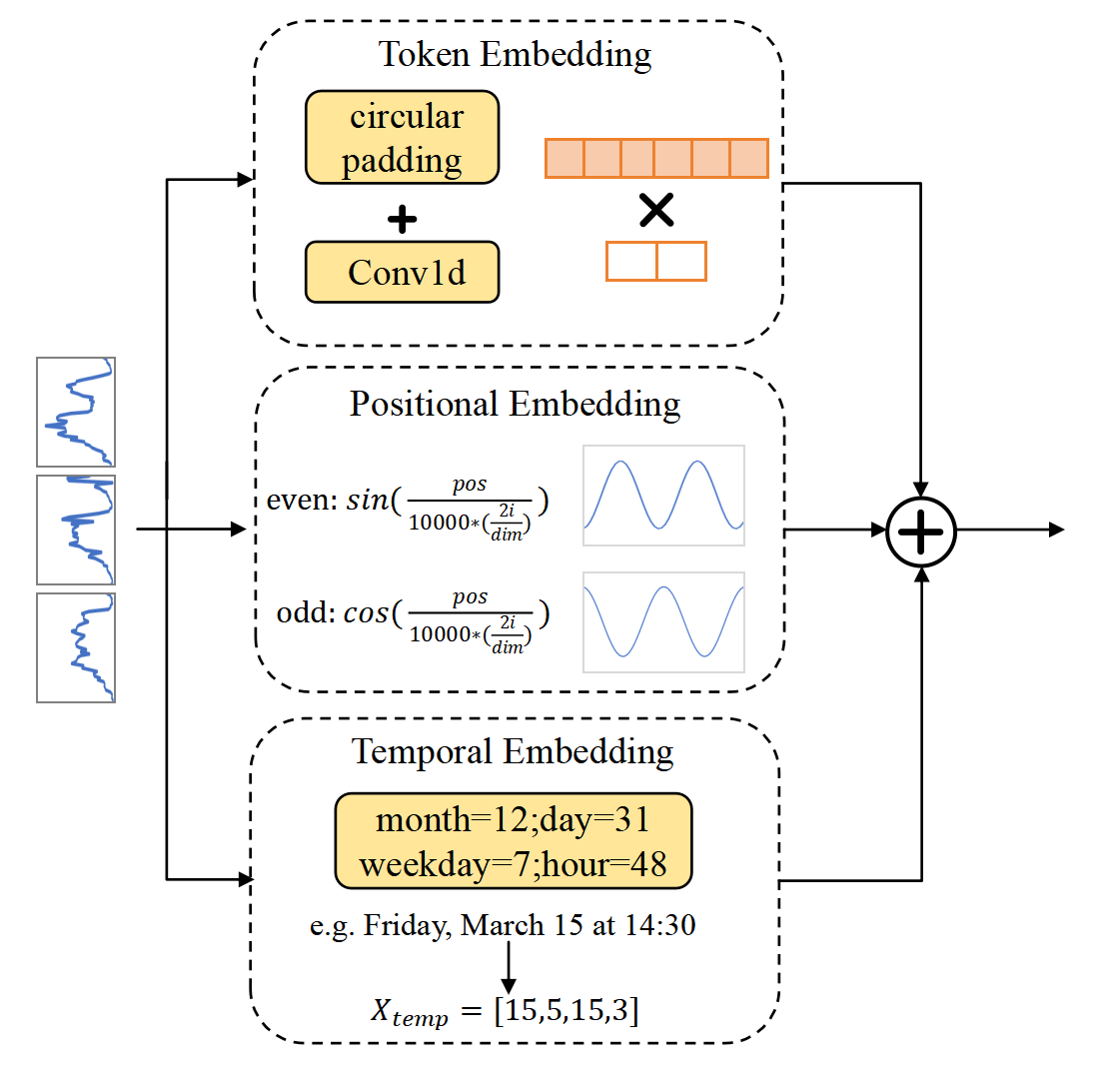}
\caption{The structure of the hybrid embedding module, which integrates three types of embeddings: token embedding, positional embedding, and temporal embedding.}
\label{fig4}
\end{figure}

The hybrid embedding module, illustrated in Fig.~\ref{fig4}, is designed to incorporate three types of embeddings: token embedding, positional embedding, and temporal embedding. Specifically, token embedding encodes the time-series data using one-dimensional convolution; positional embedding employs trigonometric functions to generate position vectors that capture relative positional relationships within the sequence; and temporal embedding transforms date information into a set of vectors to model long-term temporal patterns. By jointly leveraging these embeddings, the model is able to preserve local temporal dynamics, encode sequence order, and integrate long-term calendar effects, thereby providing a more comprehensive representation of passenger flow time series.

To capture local temporal dependencies, we apply a one-dimensional convolution on the input sequence:
\begin{align}
{{\bf{E}}_{token}} = \mathrm{Conv1D}({{\bf{\hat X}}^d};{{\bf{W}}_t})
\end{align}
where ${{\bf{W}}_t}$ denotes the convolution kernel parameters. This operation extracts short-term patterns and local fluctuations from passenger flow data.

Since transformer is order-invariant, we incorporate positional information using a trigonometric function–based encoding \cite{vaswani2017attention}:

\begin{align}
\begin{array}{l}
\mathbf{E}_{pos}(pos,2i) = \sin \left( \dfrac{pos}{10000^{2i/d_m}} \right) \\[1.3ex]
\mathbf{E}_{pos}(pos,2i + 1) = \cos \left( \dfrac{pos}{10000^{2i/d_m}} \right)
\end{array}
\end{align}
where $pos \in [0,L - 1]$ is the position index, $i$ is the dimension index, and ${d_m}$ is the embedding dimension. This embedding encodes relative position relationships within the sequence.

To capture long-term periodic patterns, we transform date-related temporal attributes—including hour of the day, day of the week, day of the month, and month—into embedding vectors. 
\begin{align}
\mathbf{E}_{temp} = f_{temp}(\mathrm{date}_t)
\end{align}
where $\mathrm{date}_t$ represents the temporal attributes of time step $t$, and $f_{temp}$ is a mapping function defined as:
\begin{align}
f_{temp} = \mathrm{Embed}(\mathrm{hour}_t, \mathrm{weekday}_t, \mathrm{day}_t, \mathrm{month}_t)
\end{align}

For example, the timestamp Friday, March 15 at 14:30 can be represented as the vector $[15,5,15,3]$, corresponding to “hour, weekday, day of the month, and month,” respectively.

The final embedding representation is formulated as:
\begin{align}
{{\bf{E}}^d} = {{\bf{E}}_{token}} + {{\bf{E}}_{pos}} + {{\bf{E}}_{temp}}
\end{align}
where ${{\bf{E}}^d} \in {\mathbb{R}^{L \times {d_m}}}$, $L$ denotes the length of the input sequence, and $d_m$ denotes the embedding dimension.

\subsection{Joint Temporal–Spectral Filtering Module}
\begin{figure*}[!t]
\centering
\includegraphics[width=15.36cm,height=5cm]{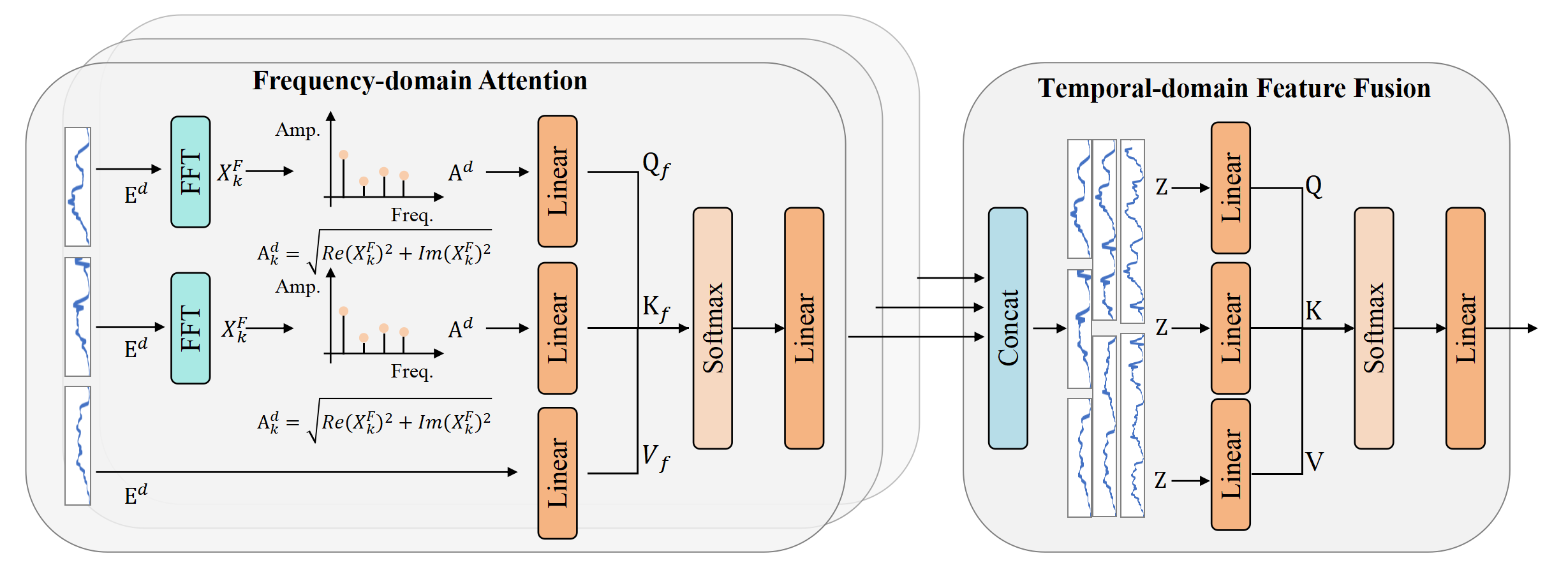}
\caption{The structure of joint temporal–spectral domain: frequency-domain attention (left) and temporal-domain feature filtering and fusion (right).}
\label{fig5}
\end{figure*}

To better capture the volatility characteristics of airport passenger flow data, we propose a joint temporal–spectral filtering modeling approach. In the frequency domain, features are extracted across multiple frequency scales, while in the time domain, filtering and enhancement are performed on the extracted features. As illustrated in Fig.~\ref{fig5}, the framework consists of two main components: a frequency-domain attention block and a temporal-domain feature fusion block. Specifically, frequency-domain attention is first applied to features at different scales to extract frequency components, followed by temporal-domain filtering and fusion to integrate multi-scale features. This enables the model to capture both the periodic patterns and short-term fluctuations of passenger flow data.

To better capture the volatility characteristics of airport passenger flow, we first conduct feature extraction in the frequency domain. Specifically, as shown on the left of Fig.~\ref{fig5}, each scale’s discrete time series is transformed using the Fast Fourier Transform (FFT). FFT efficiently implements the Discrete Fourier Transform (DFT), which converts a sequence from the time domain to the complex frequency domain. The DFT is defined as:
\begin{align}
X_k^F = \sum\nolimits_{n = 0}^{L - 1} {E_n^d} {e^{ - j2\pi kn/L}},k = 0,1, \ldots ,f
\end{align}
where ${\mathbf{E}^d}$ denotes the input sequence, $f$ is typically set to $\lfloor L/2 \rfloor$, representing the Nyquist frequency, $n$ is the time index, and $j = \sqrt{-1}$ denotes the imaginary unit. The frequency-domain representation is ${\bf{X}}^F = [X_0^F, \ldots, X_f^F] \in \mathbb{C}^{L \times d_m}$.

We adopt the amplitude of frequency components as feature representations to capture the volatility characteristics of the sequence. The amplitude is defined as:
\begin{align}
{\bf{A}}_k^d = \sqrt{ (Re(X_k^F))^2 + (Im(X_k^F))^2 }
\end{align}
where ${\bf{A}^d_k}\in {\mathbb{R}^{L \times {d_m}}}$, $Re(X_k^F)$ and $Im(X_k^F)$ denote the real and imaginary parts, respectively. By concatenating all frequency components, we obtain the amplitude feature matrix ${{\bf{A}}^d} = [{\bf{A}}_0^d,{\bf{A}}_1^d, \ldots ,{\bf{A}}_f^d] \in {\mathbb{R}^{L \times (f + 1){d_m}}}$ 

To incorporate these spectral representations into the attention mechanism, we replace the conventional linear projection with a frequency projection for generating queries and keys:
\begin{align}
{\bf{Q}}_f^d = {\bf{A}}^d {\bf{W}}_Q^f, \quad
{\bf{K}}_f^d = {\bf{A}}^d {\bf{W}}_K^f, \quad
{\bf{V}}_f^d = {\bf{E}}^d {\bf{W}}_V^f
\end{align}
where ${\bf{W}}_Q^f,{\bf{W}}_K^f,{\bf{W}}_V^f \in {\mathbb{R}^{(f+1){d_m} \times {d_m}}}$ are learnable projection matrices. This design enables attention to be computed directly in the spectral latent space:
\begin{align}
\mathbf{Z}_f^d = \mathrm{Softmax}\left( \frac{{\mathbf{Q}_f^d {\mathbf{K}_f^d}^{\top}}}{\sqrt{d_k}} \right) \mathbf{V}_f^d
\end{align}
with ${{\bf{Z}}_f^d} \in {\mathbb{R}^{L \times {d_m}}}$. By keeping the ${{\bf{V}}_f^d}$ in the time domain, our method avoids the need for an inverse Fourier transform, thereby reducing computational overhead while effectively integrating both spectral and temporal features.

After linear layer integration, the frequency domain enhanced output is obtained:
\begin{align}
{\bf{\hat Z}}_f^d = {\bf{Z}}_f^d{{\bf{W}}_O}
\end{align}
where, ${{\bf{W}}_O} \in {\mathbb{R}^{{d_m} \times {d_m}}}$.

After obtaining the spectral representations from each scale, we proceed to integrate the multi-scale features. Specifically, ${\bf{\hat Z}}_f^d \in {\mathbb{R}^{L \times {d_m}}}$ denote the outputs of the frequency-attention blocks at the $d$-th temporal scales. We first concatenate these representations along the feature dimension:
\begin{align}
{\bf{Z}} = {\rm{concat}}({\bf{\hat Z}}_f^1;\dots;{\bf{\hat Z}}_f^D)
\end{align}
where, ${\bf{Z}} \in {\mathbb{R}^{L \times (D \cdot {d_m})}}$.

In the previous stage, frequency-domain attention was applied within each scale independently, focusing on intra-scale spectral dependencies. In this stage, we fuse the representations across all scales and apply a multi-head attention mechanism to capture cross-scale dependencies:
\begin{align}
{\bf{Q}} = {\bf{Z}} {\bf{W}}_Q, \quad
{\bf{K}} = {\bf{Z}} {\bf{W}}_K, \quad
{\bf{V}} = {\bf{Z}} {\bf{W}}_V
\end{align}
where ${{\bf{W}}_Q},{{\bf{W}}_K},{{\bf{W}}_V} \in {\mathbb{R}^{(D\cdot{d_m}) \times {d_k}}}$ are learnable projection matrices. The attention output is then computed as:
\begin{align}
{\bf{\tilde Z}} = \mathrm{Softmax}(\frac{{{\bf{Q}}{{\bf{K}}^{\top}}}}{{\sqrt {{d_k}} }}){\bf{V}}
\end{align}

This multi-scale fusion enables the model to simultaneously leverage fine-grained short-term fluctuations and coarse-grained long-term dependencies, thereby enhancing its ability to capture heterogeneous temporal dynamics in airport passenger flow.

After the multi-scale attention operation, we further enhance the fused representation by passing it through a linear projection layer, which maps the aggregated features back to the target dimension. Formally, given the attention output ${\bf{\tilde Z}} \in {\mathbb{R}^{L \times {d_k}}}$, the linear layer is defined as:
\begin{align}
{\bf{Y}} = {\bf{\tilde Z W}} + b
\end{align}
where ${\bf{W}} \in {\mathbb{R}^{{d_k} \times {d_{out}}}}$ and $b \in {\mathbb{R}^{{d_{out}}}}$ are learnable parameters of the linear layer.

Finally, to obtain the forecasted passenger flow values, we apply an inverse normalization operation, which maps the outputs from the normalized training space back to the original data scale:
\begin{align}
\mathbf{\hat Y} = \text{InverseNorm}(\mathbf{Y})
\end{align}
where ${\bf{\hat Y}}$ represents the final prediction of airport passenger flow. Through this process, the model integrates multi-scale temporal-spectral information, refines it via attention and linear projection, and restores it to the real-world scale, thereby producing accurate forecasts.

\section{EXPERIMENTS}
\subsection{Datasets}
In this section, we evaluate the predictive performance of the proposed model using passenger flow data from Beijing Capital International Airport (ZBAA). Established in 1958, ZBAA has been in operation for more than 60 years. Its annual passenger throughput grew from 1.03 million in 1978 to 100 million in 2019, establishing it as one of the busiest international airports in China. The large flight volume and high operational complexity of ZBAA provide a representative and challenging dataset for model validation.

The passenger flow data were collected through the Airport Collaborative Decision Making (A-CDM) system. However, data loss occurred due to interface disconnections and camera position changes during the passenger counting process. To address this, we performed preprocessing on the historical ridership data, which included data cleaning and deletion. Large missing segments were directly removed from the dataset, while small gaps were smoothed to minimize their impact on model performance.

The raw data were recorded at the frequency of flight landings, resulting in an excessively dense time series that was not suitable for learning temporal dependencies. To address this issue, the passenger flow was aggregated into 30-minute intervals. The model was trained to predict airport passenger flow at 30-minute intervals, with forecast horizons ranging from 1 to 12 hours. Accordingly, each data point represents the total number of passengers arriving at the airport within a 30-minute window.

The dataset spans the period from January 12, 2023, to March 6, 2024, comprising 20,148 raw samples. After preprocessing, 6.8$\%$ of the data was removed, resulting in 18,766 valid samples. To eliminate scale differences in the data, Z-score normalization was applied to all original records:
\begin{align}
{Z_i} = \frac{{{X_i} - \mu }}{\sigma }
\end{align}
where ${X_i}$ is the original data sample, $\mu $ is the mean of the data, and $\sigma $ is the standard deviation of the data.

\subsection{Experimental setup}
The prediction task is formulated as predicting passenger flow over a future interval $(t+n)$ based on past passenger flow over the interval $(t-m)$. In this study, $n$ is defined within the range of 1 to 24 hours, while $m$ is defined within the range of 1 to 12 hours. At each time step $t$, the past $m$ observations and the subsequent $n$ observations constitute a single data sample. For all experiments, 70$\%$ of the dataset (13,136 samples) is used for training, 10$\%$ (1,877 samples) for validation, and the remaining 20$\%$ (3,753 samples) for testing.

The hyperparameters of the proposed model are summarized in Table~\ref{tab:model_parameters}. For fair comparison, the baseline models adopt consistent settings for key parameters, including the learning rate, the number of training epochs, and the number of hidden neurons in the linear layer, as specified in Table~\ref{tab:model_parameters}. The hyperparameters in this paper are either aligned with those reported in existing studies or selected via cross-validation to achieve optimal performance.

\begin{table}[htbp]
\centering
\fontsize{8}{11}
\renewcommand{\arraystretch}{1.4}
\caption{Model Parameters.}
\label{tab:model_parameters}
\begin{tabular}{c|lc}  
\hline
\textbf{Parameter} & \textbf{Description} & \textbf{Value} \\
\hline
patch\_num& Number of deformable partitioning modules& 3\\ 
\hline
e\_layers & Encoder layers & 4\\ 
\hline
n\_heads  & Number of multi-head attention  & 8 \\ 
\hline
d\_ff  & Number of hidden neurons in the linear layer & 256 \\ 
\hline
activation & Activation function type & Gelu  \\ 
\hline
patience& Patience for early stopping     & 40  \\
\hline
loss  & Loss function & MSE  \\ 
\hline
learning\_rate & Learning\_rate & 0.001\\ 
\hline
batch\_size & Batch size & 64 \\ 
\hline
\end{tabular}
\end{table}

\subsection{Baseline Methods and Evaluation Metrics}
We compared the performance of the proposed DTSFormer with seven representative baseline methods that have been widely adopted in airport passenger flow prediction: three classical machine learning models (SVR, BP, and RF), two recurrent neural network models (LSTM and GRU), and two patch-based Transformer models (PatchTST and Pathformer).
\begin{itemize}
    \item \textbf{SVR} \cite{hearst1998support}: Support Vector Regression is a supervised learning algorithm based on the principles of support vector machines. SVR maps the input data into a high-dimensional feature space using kernel functions and seeks a regression function that deviates from the actual observations by no more than a predefined margin, while maintaining model complexity as low as possible. By maximizing the margin and introducing slack variables, SVR achieves robust generalization and is particularly effective for modeling nonlinear relationships.
\end{itemize}
\begin{itemize}
    \item \textbf{RF} \cite{breiman2001random}: Random Forest is a non-parametric ensemble method composed of multiple decision tree learners. Each tree is defined by a hierarchical structure of decision nodes, where each node determines the branching direction based on feature evaluation. The quality of a split is measured using an impurity function.
\end{itemize}
\begin{itemize}
    \item \textbf{BP} \cite{rumelhart1986learning}: Back Propagation is a multi-layer feedforward neural network trained by the back-propagation algorithm. It is a classical nonlinear learning algorithm and has been widely applied to time-series prediction.
\end{itemize}
\begin{itemize}
    \item \textbf{LSTM} \cite{hochreiter1997long}: Long Short-Term Memory is a special type of recurrent neural network designed to mitigate the vanishing and exploding gradient problems when modeling long sequences. It captures long-term temporal dependencies by introducing memory cells and gating mechanisms.
\end{itemize}
\begin{itemize}
    \item \textbf{GRU} \cite{chung2014empirical}: Gated Recurrent Unit is closely related to LSTM but simplifies its structure by reducing three gates to two. This modification decreases the number of parameters while maintaining comparable performance in capturing long-term dependencies and alleviating gradient issues.
\end{itemize}
\begin{itemize}
    \item \textbf{PatchTST} \cite{nie2022time}: PatchTST divides the time series into subsequence-level patches, which are then used as input tokens for the Transformer. This design significantly improves long-term prediction accuracy compared with state-of-the-art Transformer-based models.
\end{itemize}
\begin{itemize}
    \item \textbf{Pathformer} \cite{chen2024pathformer}: Pathformer incorporates temporal resolution and temporal distance for multi-scale modeling. It employs patches of varying sizes to partition the time series at different temporal resolutions, thereby achieving a preliminary adaptive modeling capability.
\end{itemize}

The predictive performance of all models is evaluated for horizons of up to 12 hours ahead. To ensure fair comparisons, we adopt three widely used evaluation metrics: mean squared error (MSE), mean absolute error (MAE), and mean absolute percentage error (MAPE). MSE quantifies the average squared difference between predictions and actual values and is highly sensitive to outliers, with lower values indicating smaller overall prediction errors. MAE measures the average absolute difference, providing a more robust evaluation in the presence of outliers, since errors are not amplified by squaring. MAPE evaluates the prediction accuracy in relative terms, normalizing the error by the actual values. Collectively, these three metrics offer a comprehensive assessment of model accuracy. The formulas are as follows:
\begin{align}
\text{MSE} &= \frac{1}{a} \sum_{t=1}^{a} (\hat y_t - y_t)^2 
\end{align}
\begin{align}
\text{MAE} &= \frac{1}{a} \sum_{t=1}^{a} \left| \hat y_t - y_t \right|
\end{align}
\begin{align}
\text{MAPE} &= \frac{100\%}{a} \sum_{t=1}^{a} \left| \frac{\hat y_t - y_t}{y_t} \right|
\end{align}
where ${\hat y_t}$ is the predicted airport flow for observation $t$, ${y_t}$ is the real airport flow for observation $t$, and $a$ is the total number of observations in the test set.

\subsection{Results and Analysis}
Table~\ref{tab:forecasting_results} presents a comparison of the performance of DTSFormer against other baseline methods. The forecasting horizon ranges from 0 to 24 time steps (equivalent to 0–12 hours ahead, with a 30-minute granularity). For clarity, we report representative results at short-term, medium-term, and long-term horizons. Since longer-term prediction is of greater practical importance for airport operations, we primarily focus on the long-term forecasting performance. In Table~\ref{tab:forecasting_results}, the best results are highlighted in bold, while the second-best results are underlined.
\begin{table*}[htbp]
\centering
\caption{Airport passenger flow forecast results.}
\label{tab:forecasting_results}
\fontsize{8}{11}
\renewcommand{\arraystretch}{1.35}
\setlength{\tabcolsep}{6pt}   
\begin{tabular}{ccccccccccc}
\toprule
\textbf{Look ahead time} & \textbf{Metric} & \textbf{SVR} & \textbf{RF} & \textbf{BP} & \textbf{LSTM} & \textbf{GRU} & \textbf{PatchTST} & \textbf{Pathformer} & \textbf{DTSFormer} \\

\midrule
\multirow{3}{*}{1h}
& MSE $\downarrow$  	&48837.23	&48611.94 & 48565.65&	44656.01	&43374.74	&\underline{34271.07}&	49082.26 &	\textbf {6646.36} \\
& MAE $\downarrow$  &171.83&	154.07&	161.35	&149.92&	157.33	&\underline{129.76} &161.59& 	\textbf{56.02}  \\
& MAPE $\downarrow$ & 19.72\%& 	17.69\%& 17.87\%	& \underline{15.55\%}	& 17.40\%& 	15.92\%	& 19.36\%& \textbf{9.20\% }\\
\midrule

\multirow{3}{*}{3h}
& MSE $\downarrow$  & 55935.84&	47034.71&	60756.22	&47835.27&	39340.07&	\underline{22947.62}&	52355.22&	\textbf{4029.34}  \\
& MAE $\downarrow$ &178.48&	155.99&	175.67&	154.96&	148.41	&\underline{110.41}	&164.23&	\textbf{47.19}  \\
& MAPE $\downarrow$ & 25.39\%&	17.95\%	&20.36\%	&17.94\%&	16.85\%	&\underline{13.36\%}&	19.85\%	&\textbf{8.42\%} \\
\midrule

\multirow{3}{*}{6h}
& MSE $\downarrow$  & 50508.15	&49462.83	&56798.38&	40502.71&	40071.2	&\underline{14720.01}&	54348.94	&\textbf{3350.02}  \\
& MAE $\downarrow$  & 173.46	&157.32&	175.54	&146.38&	152.24&	\underline{93.66}	&166.11	&\textbf{44.73}  \\
& MAPE $\downarrow$& 25.63\%	&17.43\%	&20.53\%&	16.27\%	&17.71\%&\underline{12.01\%}&	18.86\%	&\textbf{7.73\%} \\
\midrule

\multirow{3}{*}{9h}
& MSE $\downarrow$ & 57120.99	&54056.98&	69830.59&	41599.27	&38407.75&	\underline{14978.75}	&54944.95	&\textbf{3681.72}  \\
& MAE $\downarrow$ & 181.5	&161.64&	189.07	&148.14	&146.62&	\underline{95.2}&	169.88	&\textbf{46.41}  \\
& MAPE $\downarrow$& 24.12\%	&18.15\%	&24.23\%&	16.11\%&	17.65\%&\underline{	11.69\%}	&19.90\%&	\textbf{7.28\%} \\
\midrule

\multirow{3}{*}{12h}
& MSE $\downarrow$  & 66133.93	&52727.81	&59734.77&	40574.27&	48749.87&\underline{22204.54}	&54199.48	&\textbf{10294.53}  \\
& MAE $\downarrow$  &193.26	&162.59	&179.3	&147.83&	164.8	&\underline{113.9}	&167.69&	\textbf{69.35}  \\
& MAPE $\downarrow$&27.11\%	&18.46\%	&21.29\%	&16.76\%	&19.89\%&	\underline{13.58\%}	&20.08\%	&\textbf{9.35\%} \\
\bottomrule
\end{tabular}
\end{table*}

\textit{1) Performance Comparison:} As shown in Table~\ref{tab:forecasting_results}, the proposed DTSFormer consistently outperforms most baseline models across all forecasting horizons. The average MAE of DTSFormer is approximately 49.47, indicating that the predicted passenger flow deviates by only about 50 persons per 30-minute interval. This deviation is relatively small compared with the average number of passengers per 30-minute interval at ZBAA (i.e., 1,135), and thus is practically meaningful for supporting ground flow management. Compared with the second-best baseline (PatchTST), DTSFormer achieves a 6.2\% reduction in MAPE, demonstrating its superior suitability for airport passenger flow forecasting.

\begin{figure*}[htbp]
\centering
\includegraphics[width=17cm,height=12cm]{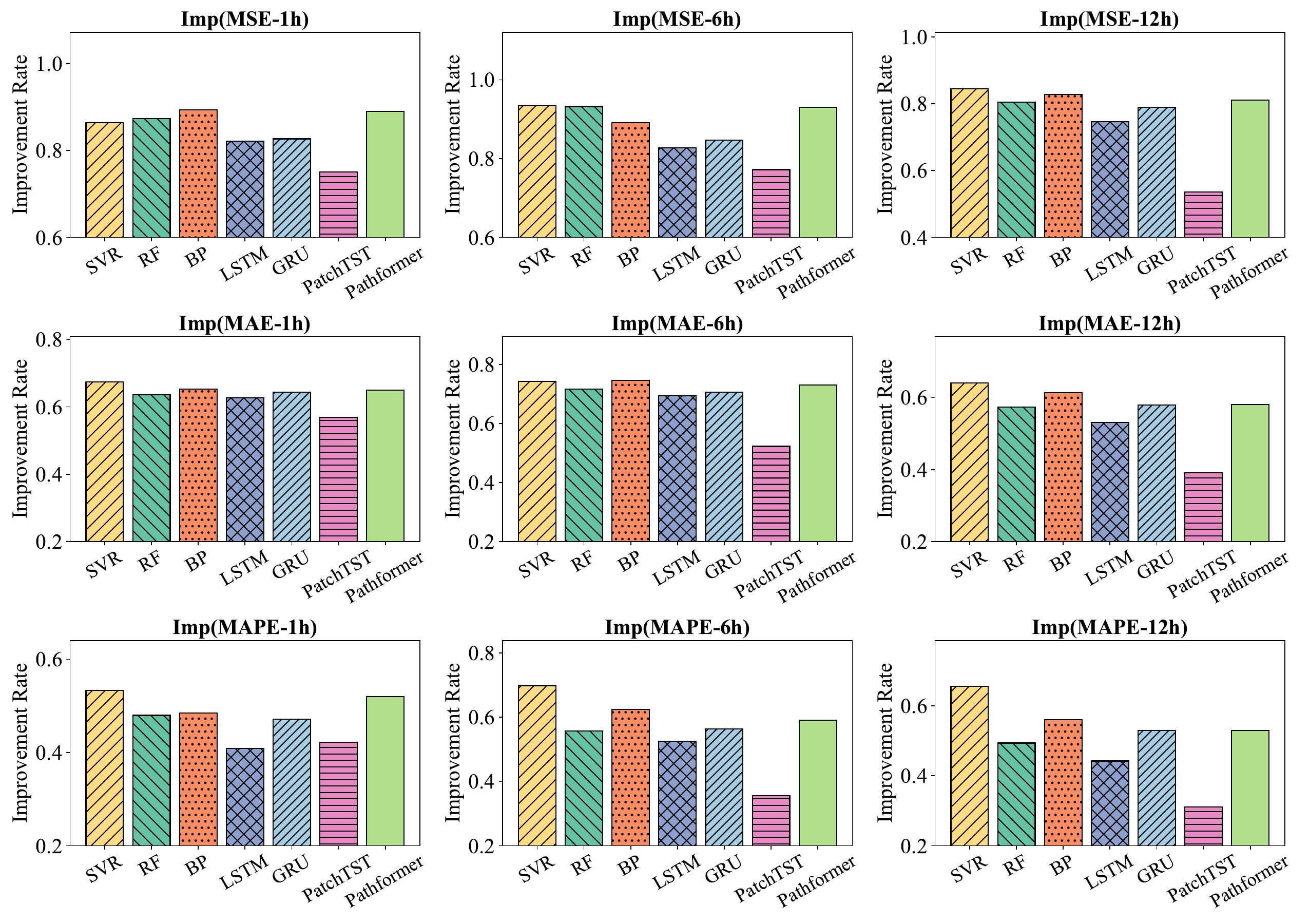}
\caption{The improvement rates of different algorithms are evaluated in terms of MSE, MAE, and MAPE across three prediction horizons (1, 6, and 12 hours). The rows correspond to the three metrics (MSE, MAE, and MAPE), while the columns correspond to the prediction horizons (1, 6, and 12 hours).}
\label{fig6}
\end{figure*}

Fig.~\ref{fig6} further illustrates the improvement ratio of our method over the baseline models. The improvement ratio is calculated as follows:

\begin{align}
\mathrm{Imp}(M) = \frac{\mathrm{Error}_{\mathrm{baseline}}(M) - \mathrm{Error}_{\mathrm{ours}}(M)}{\mathrm{Error}_{\mathrm{baseline}}(M)} \times 100\% 
\end{align}
where $\mathrm{Error}_{\mathrm{baseline}}(M)$ and $\mathrm{Error}_{\mathrm{ours}}(M)$ denote the error metrics (MAE, MSE, or MAPE) of the baseline model and DTSFormer, respectively, under metric $M$.

It is shown that the performances of DTSFormer are better than that of the baselines across all forecasting horizons. Particularly, on long time horizons, the advantages of DTSFormer become more evident with the increase in the prediction horizons.

\textit{2) Analysis of Baseline Algorithms}: As shown in Table~\ref{tab:forecasting_results}, the proposed DTSFormer consistently outperforms all baseline models. Among the classical machine learning approaches (SVR, RF, and BP), RF achieves better performance because, as an ensemble model, it integrates multiple learners to complete the prediction task, yielding higher robustness and accuracy than a single learner.

For recurrent neural networks, LSTM and GRU exhibit comparable performance since both are based on gated architectures. However, GRU requires fewer parameters than LSTM, which makes it more efficient. Moreover, in long-term forecasting, GRU outperforms RF, as it can capture temporal dependencies more effectively, which is crucial for predicting passenger flow over extended horizons.

Among the Transformer-based methods, PatchTST performs better than the other baseline algorithms, further confirming that patch segmentation is well-suited to the airport passenger flow forecasting problem. Although Pathformer is designed to adaptively select and fuse information from patches of predetermined lengths, its performance in airport passenger flow prediction is inferior to that of PatchTST. This indicates that simply adjusting patch size may not yield satisfactory results, as the starting positions of patches are equally critical. Focusing solely on scale adjustment can even introduce redundant information. Therefore, it is essential for the model to possess the capability of adaptively adjusting not only the patch size but also the number and starting positions of patches according to the characteristics of the data.

Compared with three classical machine learning models (average MAPE reduction of 12.66\%), two recurrent neural network models (average MAPE reduction of 8.81\%), and two patch-based Transformer models (average MAPE reduction of 4.91\%) across all forecast horizons, DTSFormer consistently achieves superior performance.

\subsection{Ablation study}
To further investigate the contribution of each component in the proposed model, we conduct ablation experiments through removal of specific modules. Three simplified variants are designed:

\begin{itemize}
    \item Fixed-Patch Variant (FP) (w/o deformable patch): This model employs fixed-length patches as input, removing the adaptive deformable partitioning mechanism.
    \item Linear-Mapping Variant (LM) (w/o spectral mapping): In this model, the frequency-domain mapping of queries and keys is replaced by a standard linear projection, thereby eliminating spectral feature extraction.
    \item Linear-Temporal Variant (LT) (w/o temporal filtering and fusion): This model replaces the temporal-domain filtering and fusion block with a simple linear mapping, aiming to evaluate the contribution of temporal fusion to overall performance.
\end{itemize}

\begin{table}[htbp]
\centering
\caption{Ablation study of the deformable temporal-spectral transformer, including three cases: (a) Fixed-Patch Variant, (b) Linear-Mapping Variant, and (c) Linear-Temporal Variant}
\label{tab: Ablation study}
\fontsize{8}{11}
\renewcommand{\arraystretch}{1.35}
\setlength{\tabcolsep}{5pt}   
\begin{tabular}{>{\centering}p{1.2cm}ccccc}
\toprule
\textbf{Look ahead time} & \multirow{2}{*}{\textbf{Metric}} & \multirow{2}{*}{\textbf{FP}} & \multirow{2}{*}{\textbf{LM}} & \multirow{2}{*}{\textbf{LT}} &\multirow{2}{*}{\textbf{DTSFormer}} \\
\midrule
\multirow{3}{*}{1h}
& MSE $\downarrow$  & 33867.25&	11938.77& 20783.73 &	\textbf{6646.36}  \\
& MAE $\downarrow$  & 136.44	 &    72.75 &	103.43 &\textbf{56.02}   \\
& MAPE $\downarrow$  & 13.20\%&	10.53\%& 11.02\%	&\textbf{9.20\%}  \\
\midrule

\multirow{3}{*}{3h}
& MSE $\downarrow$  & 31721.52 &	7917.26 &6607.86 &\textbf{4029.34}  \\
& MAE $\downarrow$  & 134.38 &	60.90 &56.58  &\textbf{47.19}  \\
& MAPE $\downarrow$ & 14.26\% &	10.01\% & 8.96\%&\textbf{8.42\%}  \\
\midrule

\multirow{3}{*}{6h}
& MSE $\downarrow$ & 28548.41 &	4885.09 &7761.21  &\textbf{3350.02}  \\
& MAE $\downarrow$ & 125.4	 &	52.56  &63.01  &\textbf{44.73}  \\
& MAPE $\downarrow$ & 13.72\% &	8.80\% & 8.74\%&\textbf{7.73\%}  \\
\midrule

\multirow{3}{*}{9h}
& MSE $\downarrow$ & 30272.34 &	8347.06 &16890.36  &\textbf{3681.72 }  \\
& MAE $\downarrow$ & 129.86 &	64.78 &90.38  &\textbf{46.41 }  \\
& MAPE $\downarrow$ & 13.42\% &	9.58\% &9.74\% &\textbf{7.28\%}  \\
\midrule

\multirow{3}{*}{12h}
& MSE $\downarrow$ & 36798.11 &	22846.97 &16830.86  &\textbf{10294.53}  \\
& MAE $\downarrow$ & 142.54 &	102.02 &92.51  &\textbf{69.35 }  \\
& MAPE $\downarrow$ & 15.98\% &	13.27\% &10.28\% &\textbf{9.35\%}  \\
\bottomrule
\end{tabular}
\end{table}
\begin{figure*}[!t]
\centering
\includegraphics[width=17cm,height=5cm]{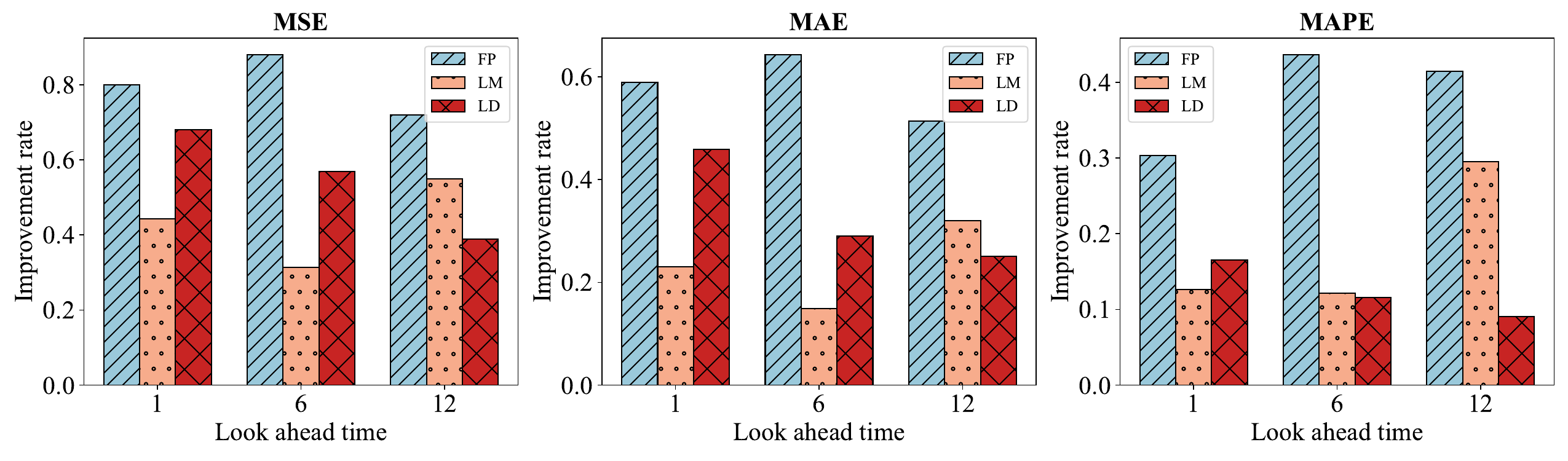}
\caption{The improvement rates of three models that are built by removing certain parts from the proposed model are evaluated in terms of MSE, MAE, and MAPE across three prediction horizons (1, 6, and 12 hours).}
\label{fig7}
\end{figure*}
Their performance is shown in Table~\ref{tab: Ablation study} and Fig.~\ref{fig7}. Removing the multi-scale deformable partitioning module yields a 5.72\% average performance drop, indicating that adaptive, data-driven patching is a primary source of accuracy gains. This module enables the model to adjust patch number, size, and offsets to capture heterogeneous temporal patterns. Likewise, using linear projection instead of spectral mapping leads to a 2.04\% decrease, confirming the necessity of explicit frequency-domain feature extraction. Similarly, disabling the time domain filter and fusion module results in a 1.81\% decrease, indicating that the frequency components extracted in the previous step are not equally important, and therefore require differentiated weighting in the temporal domain, leading to more effective utilization of spectral information and improved prediction accuracy. Together, these components enable DTSFormer to model both multi-scale temporal trends and high-frequency volatility, resulting in superior predictive performance.

\section{Further Analysis}
\subsection{Analysis of Learned Partitioning}
To further analyze the effectiveness of the multi-scale deformable partitioning module, we conducted a visualization study. The visualization allows us to observe the distribution characteristics of the segmented patches and verify that the data-driven partitioning aligns with the inherent patterns of airport passenger flow. First, we analyze the patch lengths across different scales and observe that the patch sizes are closely related to the periodic properties of the data. Next, we examine the positions of the patches and observe that they are associated with the heterogeneous trends within the data. The detailed analysis is presented as follows.

\begin{figure}[!t]
\centering
\includegraphics[width=8.5cm,height=5.5cm]{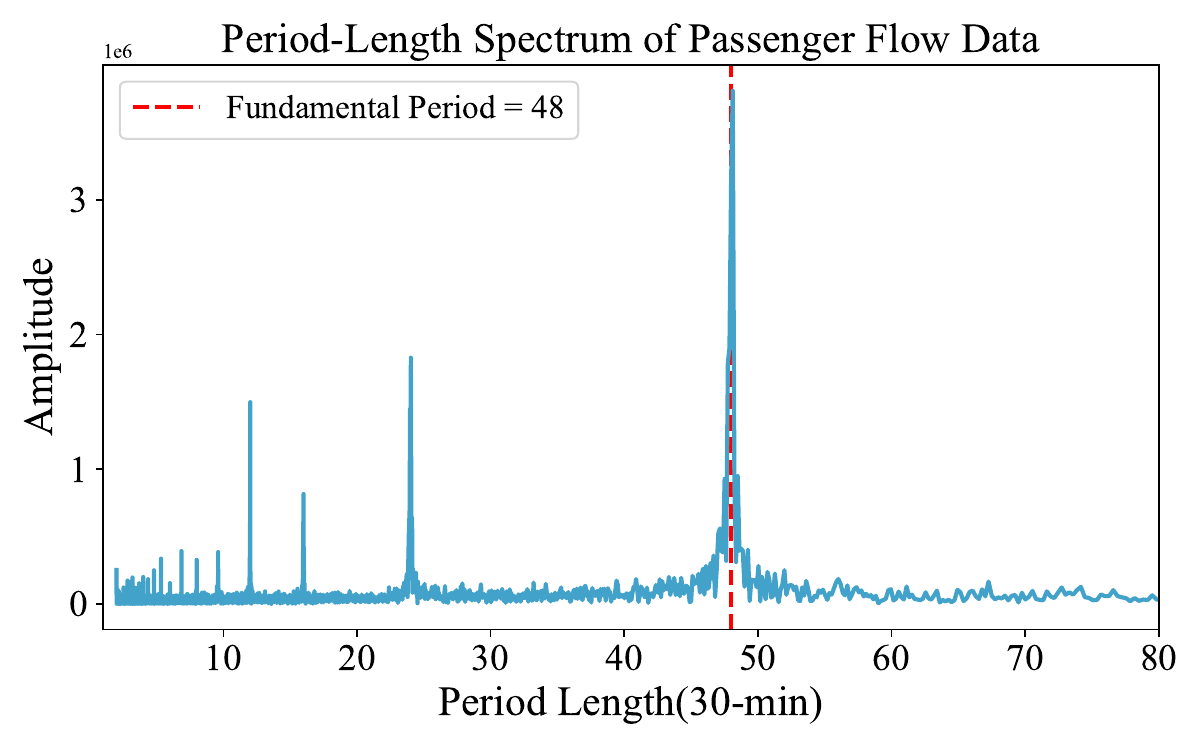}
\caption{Fourier spectrum analysis of airport passenger flow data. The x-axis denotes the period length, and the y-axis represents the amplitude obtained via the Fourier transform.}
\label{fig8}
\end{figure}

First, we analyze the patch sizes across multiple scales. After model training, we obtain the deformable partitioning results. Specifically, Layer 1 produced patches of length 3, Layer 2 produced patches of length 6, and Layer 3 produced patches of length 8. By applying the Fourier transform to the entire dataset, we identify the most significant periodic length as 48 (Fig.~\ref{fig8}). The learned patch lengths correspond to fractional multiples of this fundamental period, which facilitates the model's ability to effectively capture the underlying periodic patterns \cite{cao2007discovery}.

\begin{figure}[!t]
\centering
\includegraphics[width=9cm,height=5cm]{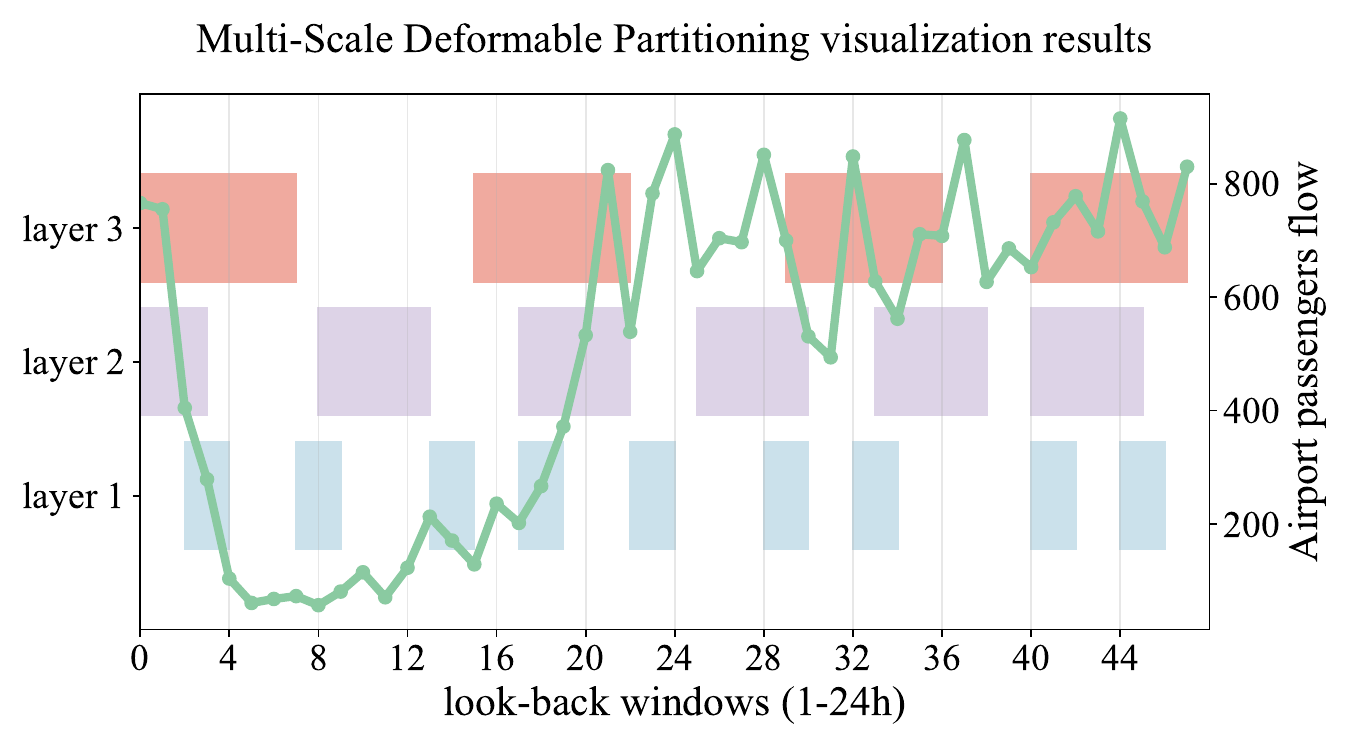}
\caption{Visualization results of multi-scale deformable partitioning. Layers 1, 2, and 3 represent the results of the three deformable layers.}
\label{fig9}
\end{figure}

Then, we examine the positions of the patches. Fig.~\ref{fig9} presents the coverage ranges of patches across different scales for the case of January 5, 2023. Each colored bar represents a patch interval, clearly revealing the distribution of patches at multiple scales. Layer 1 appears more densely and is responsible for capturing local fluctuations, whereas layer 3 covers broader intervals to model long-term dependencies. The visualization results show that the learned patch lengths differ across layers, indicating that the model adaptively partitions the data at multiple temporal scales. The partitioning is not uniform; instead, the model adaptively selects patch boundaries based on data characteristics such as peaks and turning points, highlighting its flexibility. Specifically, as illustrated in Fig.~\ref{fig9}, the patches in Layer 1 primarily emphasize short-term upward or downward movements, where the trend patterns are relatively simple and exhibit little fluctuation. In contrast, Layer 2 focuses on more complex variations, such as V-shaped or W-shaped changes, capturing both rising and falling dynamics within a short horizon. Layer 3 highlights long-term transformations that align with the actual operational patterns of the airport, including nighttime troughs as well as morning and afternoon peaks. These results demonstrate that the multi-scale partitioning module effectively disentangles heterogeneous temporal trends.

\subsection{Analysis of High-Frequency Feature Extraction}
To further evaluate the capability of our model to capture high-frequency features, we conduct a case study on the prediction results for July 31, 2023. On this day, Beijing received an average rainfall of 331 mm, which significantly impacted passenger flow at the airport. For most of the time, aircraft were circling in the air awaiting clearance, and only during intervals when the rainfall subsided did a large number of flights land simultaneously. This led to a sudden surge in passenger arrivals within a short period of time, making the high-frequency characteristics of the flow particularly prominent on that day.

\begin{figure}[!t]
\centering
\includegraphics[width=9cm,height=6cm]{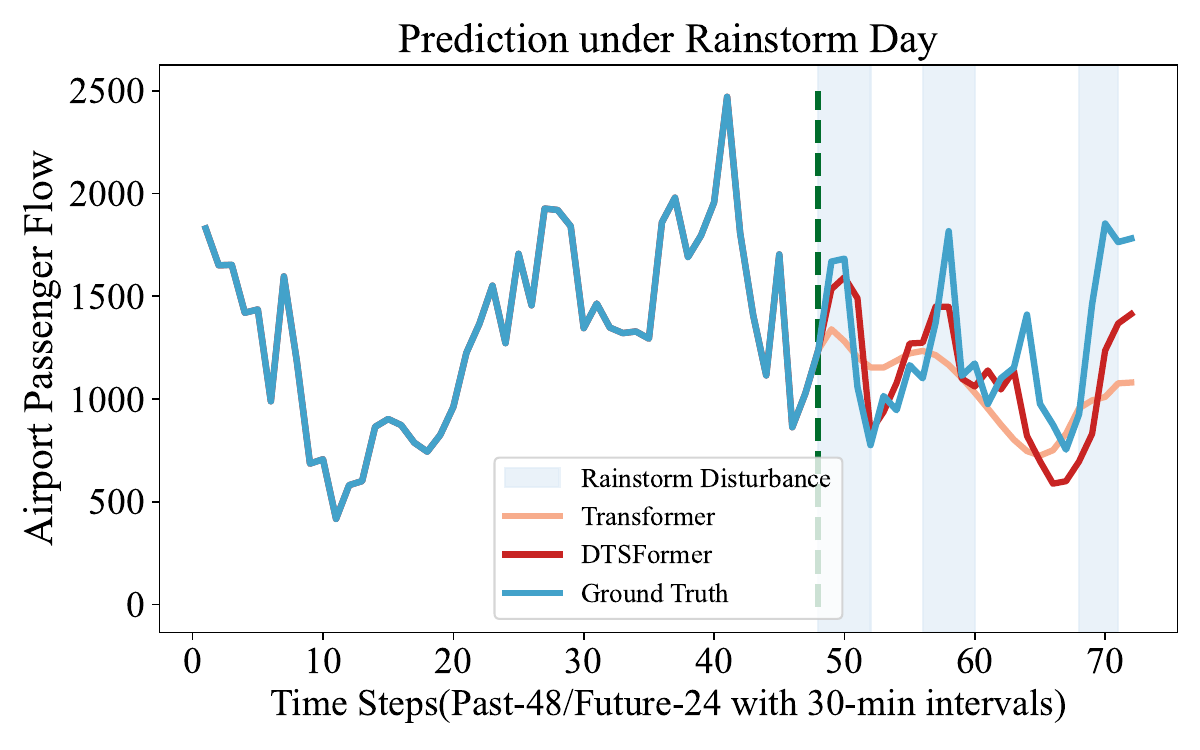}
\caption{Passenger flow prediction result under rainstorm conditions that have characteristics of distinct high-frequency.}
\label{fig10}
\end{figure}

We compare the predictions of the proposed model with those of the Transformer baseline. As shown in Fig.~\ref{fig10}, the blue-shaded area highlights the regions of rapid passenger flow fluctuations. Our model provides accurate predictions of these sudden changes, demonstrating its effectiveness in capturing high-frequency components and thereby improving the forecasting accuracy of passenger flow.

\section{CONCLUSION AND FUTURE WORK}
Accurate airport passenger flow prediction is crucial for effective airport management and resource allocation. The research objective of this paper is to propose a novel forecasting framework to address the challenges posed by heterogeneous trends and high-frequency variations in passenger flow data. The framework integrates a multi-scale deformable partitioning mechanism with a joint temporal–spectral filtering module to enhance the model's ability to extract periodic and dynamic features.
Comprehensive experiments on real-world data from Beijing Capital International Airport demonstrate that the proposed method consistently outperforms state-of-the-art prediction models in terms of MSE, MAE, and MAPE across multiple forecasting horizons. Further analysis confirms that the model effectively captures high-frequency variations, enabling more accurate and responsive passenger flow estimation.

In future work, we plan to incorporate external knowledge sources, such as weather information and air traffic control measures, to further improve the model's robustness, applicability, and scalability in complex real-world operational environments.

\bibliographystyle{IEEEtran}
\bibliography{ref} 

\newpage

\vspace{11pt}

\vspace{11pt}

\vfill

\end{document}